\documentclass[sn-mathphys-num]{sn-jnl}% Math and Physical Sciences Numbered Reference Style 
\usepackage{graphicx}%
\usepackage{multirow}%
\usepackage{amsmath,amssymb,amsfonts}%
\usepackage{amsthm}%
\usepackage{mathrsfs}%
\usepackage[title]{appendix}%
\usepackage{xcolor}%
\usepackage{textcomp}%
\usepackage{manyfoot}%
\usepackage{booktabs}%
\usepackage{algorithm}%
\usepackage{algorithmicx}%
\usepackage{algpseudocode}%
\usepackage{listings}%
\usepackage{subfigure}
\theoremstyle{thmstyleone}%
\newtheorem{theorem}{Theorem}%  meant for continuous numbers
\theoremstyle{thmstyletwo}%

\theoremstyle{thmstylethree}%

\begin{document}

\title[Article Title]{Signal Adversarial Examples Generation for Signal Detection Network via White-Box Attack}


\author[1,2]{\fnm{Dongyang} \sur{Li}}\email{easunlee1006@126.com}

\author*[1]{\fnm{Linyuan} \sur{Wang}}\email{wanglinyuanwly@163.com}

\author[1]{\fnm{Guangwei} \sur{Xiong}}\email{GuangweiX@hotmail.com}

\author[1]{\fnm{Bin} \sur{Yan}}\email{ybspace@hotmail.com}

\author[1]{\fnm{Dekui} \sur{Ma}}\email{dekuima@qq.com}

\author[2]{\fnm{Jinxian} \sur{Peng}}\email{568859101@qq.com}

\affil*[1]{\orgdiv{Information Engineering College}, \orgname{Information Engineering University}, \orgaddress{\street{Jianxue Street}, \city{Zhengzhou}, \postcode{450001}, \state{Henan Province}, \country{China}}}

\affil[2]{\orgdiv{63611 Troops}, \orgname{People's Liberation Army of China}, \orgaddress{\city{Kuerle}, \postcode{841000}, \state{Xinjiang Province}, \country{China}}}


\abstract{With the development and application of deep learning in signal detection tasks, the vulnerability of neural networks to adversarial attacks has also become a security threat to signal detection networks. This paper defines a signal adversarial examples generation model for signal detection network from the perspective of adding perturbations to the signal. The model uses the inequality relationship of L2-norm between time domain and time-frequency domain to constrain the energy of signal perturbations. Building upon this model, we propose a method for generating signal adversarial examples utilizing gradient-based attacks and Short-Time Fourier Transform. The experimental results show that under the constraint of signal perturbation energy ratio less than 3$\%$, our adversarial attack resulted in a 28.1$\%$ reduction in the mean Average Precision (mAP), a 24.7$\%$ reduction in recall, and a 30.4$\%$ reduction in precision of the signal detection network. Compared to random noise perturbation of equivalent intensity, our adversarial attack demonstrates a significant attack effect.}




\maketitle

\section{Introduction}\label{sec1}

The application of deep learning in the field of signal processing has been extensively developed\cite{RN166}, such as in signal recognition\cite{RN172, RN4}, spectrum sensing\cite{RN487, RN488}, signal detection\cite{RN177, RN176, RN491, RN174, RN173}, and signal resource allocation\cite{RN178}. Signal detection refers to the construction of detection features from broadband received signal data to effectively guide signal processing tasks such as modulation recognition, channel encoding/decoding, and radiation source individual identification, which are fundamental to signal processing. The signal detection network based on deep learning is typically constructed using the application paradigm of signal time-frequency diagrams and object detection. Signal detection networks have low requirements for signal quality, can accurately capture the timing and frequency distribution information of signals, and is capable of detecting multiple signal targets simultaneously. It features strong anti-interference, good detection effects, and high detection efficiency. Based on this application paradigm, researchers have conducted a series of studies and applications in different scenarios and frequency bands\cite{RN177, RN176, RN491, RN174, RN173}. 

Signal detection networks has shown diverse advantages in various application scenarios. However, the vulnerability of neural networks is an issue that cannot be ignored. Szegedy\cite{RN163} pointed out that neural networks generally have vulnerabilities, meaning that adding minor perturbations to input samples can affect the model's output, and even lead to incorrect results. These perturbed examples are known as adversarial examples. In signal processing field, adversarial attacks still exist. In the task of signal modulation recognition, Toshea et al.\cite{RN166} in 2016 first applied deep learning to this task, demonstrating that deep learning algorithms outperform expert features and higher-order moment methods in signal modulation recognition. Subsequently, Sadeghi et al.\cite{RN283, RN159, RN270} conducted in-depth studies on adversarial attacks against signal modulation recognition networks and found that the network is vulnerable to signal adversarial examples under various signal-to-noise ratios (SNRs). Moreover, Sadeghi and Flowers further noted that compared to Gaussian noise perturbations of the same intensity, signal adversarial examples exhibit a more significant attack effect\cite{RN218, RN270}. Additionally, adversarial attacks have been studied in spectrum sensing\cite{RN304, RN202, RN215, RN220} and signal resource allocation tasks\cite{RN315, RN221, RN490}. 

When a signal detection network is subjected to adversarial attacks, it may result in missed or false detections, hindering the correct processing of subsequent signals and potentially causing serious communication security issues. While the research specifically addressing adversarial attacks against signal detection networks is limited, it is worth noting that the field of adversarial attack studies for object detection is comparatively mature. It includes both black-box and white-box attack methods, with a variety of attack objectives such as attacks on bounding boxes\cite{RN493, RN496}, classification\cite{RN492, RN494}, and contextual semantics\cite{RN497}. Nevertheless, current adversarial attack methods on object detection networks are limited to targeting the time-frequency diagrams of signals, rather than directly generating adversarial examples for signals by adding perturbations. Moreover, compared to signal modulation recognition networks that extract features directly from signal, the complexity of time-frequency transformation makes direct gradient backpropagation difficult when generating signal adversarial examples. Therefore, gradient-based attacks targeting signal detection network require more consideration and innovative solutions.

In light of the potential threats posed by adversarial examples to signal detection network, this paper presents a model generating signal adversarial examples, based on time-frequency transformations and the inequality relationship of the L2-norm between time-frequency domain perturbations and time domain perturbations. By comparing with original clean signal examples and random noise interference under the same constraints, the effectiveness of the attack model is validated, revealing the vulnerability of signal detection network when faced with signal adversarial examples attacks.

\section{Method}\label{sec2}

This chapter primarily focuses on the generative model of signal adversarial examples for signal detection networks, as well as the method for generating adversarial examples under white-box conditions. Approaching from the perspective of adding perturbations to signals and taking into account the actual attack conditions and objectives, we delineate the signal adversarial examples generation model based on the L2-norm’s inequality relationship between the time-frequency domain and the time domain. Building upon this model, we propose a methodology for generating signal adversarial examples in white-box scenarios.

\subsection{Signal Adversarial Examples Generation Model}\label{subsec2}

This section delves into the signal adversarial examples generation model. It begins by delineating the mathematical framework of the signal detection network, establishing a foundation for generating adversarial examples. Subsequently, drawing upon research in adversarial attacks on object detection networks, it presents a definition for time-frequency diagram adversarial examples. Finally, the signal adversarial examples generation model is proposed, which enforces constraints on the signal energy through the L2-norm inequality relationship between time-frequency domain perturbations and time domain perturbations.

\subsubsection{Mathematical Framework for Signal Detection Networks}\label{subsubsec2}

The structure of signal detection network is illustrated in Figure 1, consisting of two cascaded parts: the signal time-frequency diagram conversion module and the object detection network.

\begin{figure}[h]
\centering
\includegraphics[width=0.7\columnwidth]{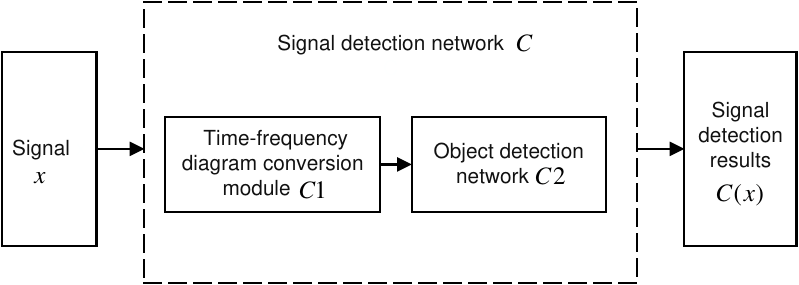}
\caption{\centering{Signal detection network structure}}\label{fig1}
\end{figure}

The time-frequency conversion module $C1$ processes signal data through Short-Time Fourier Transform (STFT) to produce a grayscale time-frequency diagram. It transforms the signal into a complex matrix representing amplitude and phase, converts amplitudes to decibels, and maps them onto a diagram where pixel value indicates signal magnitude over time and frequency.

The object detection network $C2$ is cascaded after the time-frequency diagram conversion, taking the transformed signal time-frequency diagram as its input. The output consists of the detected signal categories and positional information. In the positional information, the horizontal coordinate corresponds to the time information of the signal target in the time-frequency diagram, and the vertical coordinate corresponds to the frequency information of the signal target.

The signal detection network, constructed by cascading the time-frequency diagram conversion module $C1$ and the object detection network $C2$, is capable of directly reading signal data and outputting the detected signal categories, frequencies, and time information, thereby achieving end-to-end detection. The time-frequency diagram and detection results are shown in Figure 2. The horizontal axis represents time dimension, and the vertical axis denotes frequency dimension. The elongated white bars depicted in the figure correspond to the signal objectives. As illustrated, in the task of signal detection, a segment of signal data encompasses numerous signal targets distributed across various frequencies and time intervals.

\begin{figure}[htbp]
\centering
\subfigure[]
{
    \begin{minipage}[b]{.45\linewidth}
        \centering
        \includegraphics[scale=0.8]{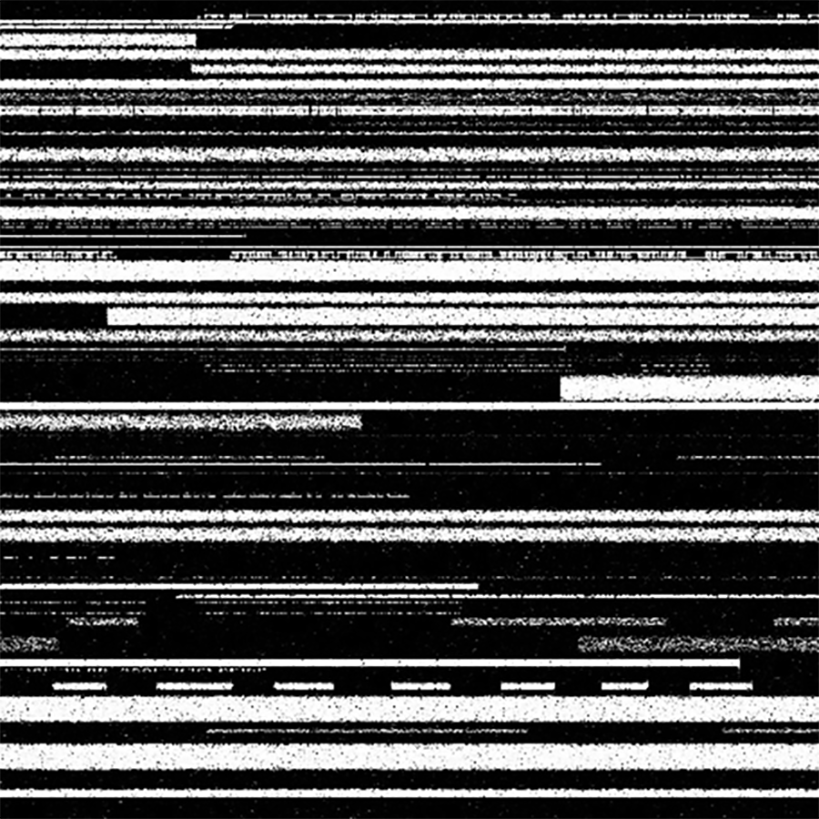}
    \end{minipage}
}
\subfigure[]
{
 	\begin{minipage}[b]{.45\linewidth}
        \centering
        \includegraphics[scale=0.8]{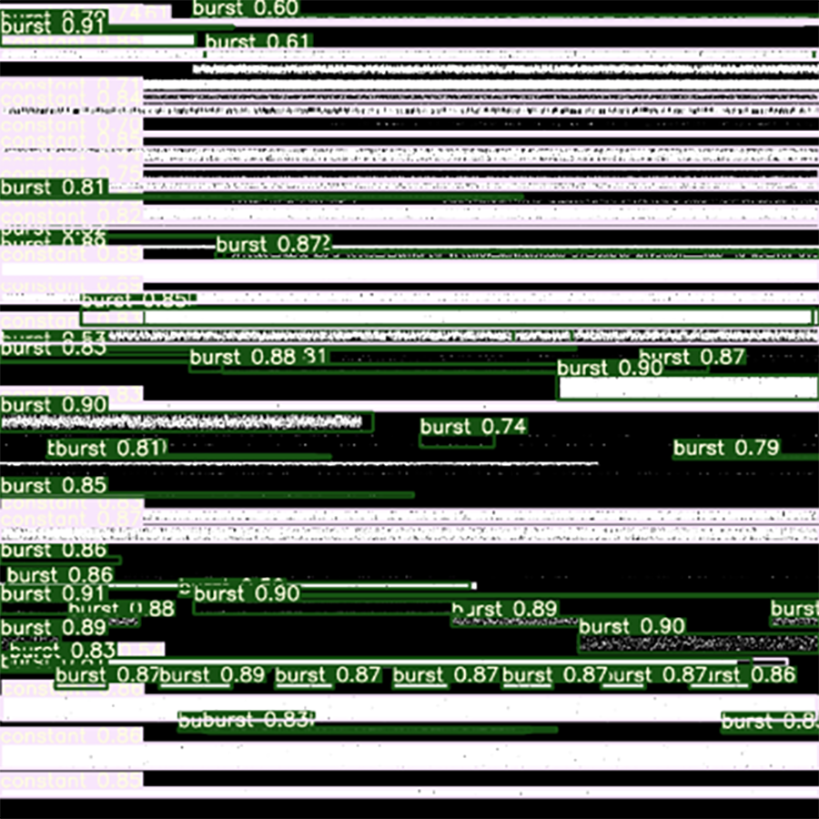}
    \end{minipage}
}
\caption{\centering{Signal time-frequency diagram and detection results}}\label{fig2}
\end{figure}

Based on the structure of signal detection network, we provide the mathematical expression of the network as follows:
\begin{equation}
f(x ; \theta):=f_2\left(f_1(x) ; \theta\right)
\end{equation}

Let the time-frequency diagram conversion module $C1$ be denoted as $\mathop{y=f}_{1}(x)$, with the signal sample input as $x$. After processing by $C1$, it is converted into a time-frequency diagram $y$, where $y$ primarily contains the amplitude information within the signal's time-frequency domain. Let the object detection network $C2$ be denoted as $\mathop{f}_{2}(y\ ;\theta )$, where $\theta $ represents the model parameters of the detection network. The detection network ${{f}_{2}}$ analyzes the generated time-frequency diagram $y$, and the output includes the category, position, and dimensions of the signal target.

\subsubsection{Adversarial Examples for Signal Time-frequency Diagram}\label{subsubsec2}

The amplitude information in the time-frequency domain $\mathop{y=f}_{1}(x)$ corresponds to the output of the signal $x$ through the time-frequency diagram conversion module $C1$. Informed by the current body of research on adversarial examples within object detection networks, we can implement adversarial attacks targeting the time-frequency diagram $\mathop{y=f}_{1}(x)$ of signals. 

In practical attack scenarios, the goal is to add perturbations to the signal such that signal detection network miss a large number of signal targets. Consequently, we define the time-frequency domain signal adversarial examples which minimize the number of detected targets, under the constraint that the perturbation’s L2-norm is less than a certain proportion of the original signal's time-frequency domain L2-norm as follows:
\begin{equation}
\begin{matrix}
  \min \text{  }{{f}_{2}}({{y}_{r}}+{{\beta }_{r}}\ ;\theta ) \\ 
  s.t.\quad {{y}_{r}}=\mathop{f}_{1}({{x}_{r}}) \\ 
  ||{{\beta }_{r}}|{{|}_{2}}\le ||{{y}_{r}}|{{|}_{2}}*\alpha  \\ 
\end{matrix}
\end{equation}
${{x}_{r}}$ is original input signal, $||{{\beta }_{r}}|{{|}_{2}}$ is the L2-norm of the time-frequency domain adversarial perturbation, $||{{y}_{r}}|{{|}_{2}}$ is the L2-norm of the time-frequency domain signal sample, $\alpha $ is the maximum ratio value of $||{{\beta }_{r}}|{{|}_{2}}$ compared to $||{{y}_{r}}|{{|}_{2}}$ and the time-frequency domain signal adversarial example is $y_{r}^{*}={{y}_{r}}+{{\beta }_{r}}$.

Given that the attack targets the time-frequency diagram, we could adopt attack methodologies to object detection networks. The primary objective of the attack is to induce the substantial disappearance of signal targets. Drawing from the approach delineated in TOG\cite{RN493}, we define the loss function of time-frequency domain signal adversarial examples as follows:
\begin{equation}
\begin{matrix}
{{L}_{obj}}(\tilde{y};O,\theta )=\sum\limits_{i=1}^{S}{\left[ {{1}_{i}}{{l}_{BCE}}(1,{{{\hat{C}}}_{i}}) \right]}\\
{{L}_{noobj}}(\tilde{y};O,\theta )=\sum\limits_{i=1}^{S}{\left[ (1-{{1}_{i}}){{l}_{BCE}}(0,{{{\hat{C}}}_{i}}) \right]}\\
L({y}';\varnothing ,\theta )={{L}_{obj}}({y}';\varnothing ,W)+\lambda {{L}_{noobj}}({y}';\varnothing ,W)\\
\end{matrix}
\end{equation}
Let $S$ be the total number of candidate boxes, where $i$ is the candidate box’s number, and ${{\hat{C}}_{i}}$ is the existence probability of this box. The information set predicted for each box ${{o}_{i}}$ includes the center position, length, width and classification confidence of the box, which is defined as $O=\left\{ {{o}_{i}}|{{1}_{i}}=1,1\le i\le S \right\}$, where ${{o}_{i}}=\left( b_{i}^{x},b_{i}^{y},b_{i}^{W},b_{i}^{H},{{p}_{i}} \right)$. Assume that there is an object contained within box 1, the judgment item ${{1}_{1}}$ is 1, then as the predicted value of ${{\hat{C}}_{1}}$ becomes larger, the value of ${{L}_{obj}}$ becomes smaller. Conversely, when box 2 contains no object, and the judgment item ${{1}_{2}}$ is 0, then as ${{\hat{C}}_{2}}$ becomes smaller, ${{L}_{noobj}}$ also becomes smaller. 

Our attack goal is to minimize the signal targets’ numbers under the perturbation L2-norm constraint. Therefore, we can transform the problem and define the loss function as $L({y}';\varnothing ,\theta )$. 

By using this loss function, we can calculate the gradient direction of signal targets’ disappearance, thereby generating adversarial examples of the time-frequency diagram.

\subsubsection{Signal Adversarial Examples Generation Model}\label{subsubsec2}

Building upon the definition of adversarial examples for signal time-frequency diagrams and the adversarial attack methods for object detection networks, we could achieve adversarial attacks on time-frequency diagrams. However, we are still unable to directly impose perturbations on signals to generate signal adversarial examples. A critical issue is controlling the magnitude of perturbations added to the time domain adversarial examples to ensure they remain imperceptible. Fortunately, we investigated the relationship between the L2-norm of time-frequency domain perturbations and time domain perturbations, and conclude that controlling the energy of perturbations added to the time domain can be achieved by constraining the L2-norm of time-frequency domain perturbations. This insight has led to the proposal of a generative model for signal adversarial examples.

Next, we will build the relationship between the perturbations in the time-frequency domain and the perturbations in the time domain with respect to their L2-norm:

\begin{figure}[h]
\centering
\includegraphics[width=0.7\columnwidth]{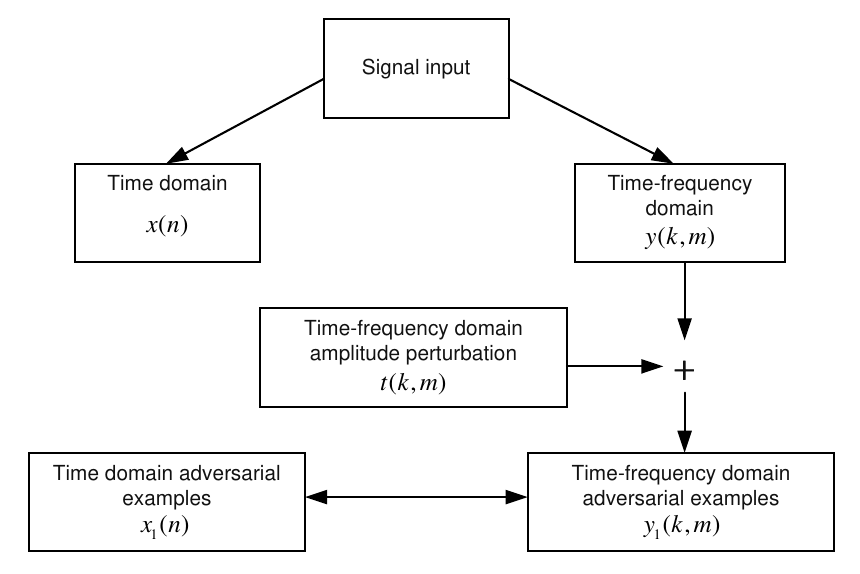}
\caption{\centering{Schematic diagram of signal detection network adversarial examples' time-frequency conversion}}\label{fig3}
\end{figure}

As shown in Figure 3, assume that the input signal sequence is $x[n]$, after short-time Fourier transform, time-frequency matrix $y(k,m)=\mathop{f}_{1}(x[n])$ is obtained. The transformation formula is as follows: 
\begin{equation}
y(k,m)=\sum\limits_{n=0}^{N-1}{x[n]w[n-mR]{{e}^{\frac{-j2\pi kn}{N}}}}
\end{equation}
where $k$ is the frequency index, $m$ is the time index, $R$ is the moving step size of the time window $w$, and $y(k,m)$ is a complex matrix containing the amplitude and phase information of signal $x[n]$. Adding perturbation $t(k,m)$ to the amplitude value of $y(k,m)$, the resulting new signal time-frequency matrix ${{y}_{1}}(k,m)$ is:
\begin{equation}
{{y}_{1}}(k,m)=\left( \left| y(k,m) \right|+t(k,m) \right){{e}^{j\angle y(k,m)}}
\end{equation}
The L2-norm of the perturbation added to the amplitude in the time-frequency domain is given by: 
\begin{equation}
||{{\beta }_{r}}|{{|}_{2}}=\sqrt{\sum\limits_{k=0}^{N-1}{\sum\limits_{m=-\infty }^{\infty }{{{t}^{2}}(k,m)}}}
\end{equation}
Here, $t(k,m)$ represents the perturbation adding to time-frequency matrix. This equation quantifies the magnitude of the perturbation in the time-frequency domain. 

The time-frequency domain signal adversarial example ${{y}_{1}}(k,m)$ is restored to the time domain ${{x}_{1}}[n]$ through the inverse short-time Fourier transform. Then the corresponding L2-norm of time domain perturbation ${{\delta }_{r}}$ is:
\begin{equation}
||{{\delta }_{r}}|{{|}_{2}}={{\left\| {{x}_{1}}-x \right\|}_{2}}
\end{equation}

Based on the Fourier transform and the inequality relationship, we proved the following theorem:

\begin{theorem}\label{thm1}
Suppose ${{\delta }_{r}}$ is the perturbation vector added to the signal in the time domain, ${{\beta }_{r}}$ is the perturbation matrix added to the magnitude of the signal in the time-frequency domain, and $N$ is the length of the time window, the inequality relationship between the two is as follows:
\begin{equation}
||{{\delta }_{r}}|{{|}_{2}}\le \sqrt{\frac{3}{N}}||{{\beta }_{r}}|{{|}_{2}}
\end{equation}
\end{theorem}

This theorem is based on Short-time Fourier Transform and the inequality relationship of vector dot products. The proof is contained in Appendix A. Consequently, we can indirectly constrain the L2-norm of the perturbation in the time domain by restricting the L2-norm of the amplitude perturbation in the time-frequency domain, thereby controlling the magnitude of perturbation. 

In addition, unlike signal modulation recognition, signal detection tasks are characterized by multiple signal targets, long signal data duration, and wide frequency bands. It is not feasible to directly use the signal-to-noise ratios (SNRs) to depict the magnitude of the perturbation; therefore, the L2-norm is used to represent the energy, characterizing the imperceptibility of the signal perturbation, which is consistent with the actual attack scenario.

In practical attack scenarios, the goal is to add perturbations to the signal such that signal detection network miss a large number of signal targets. Consequently, based on the Theorem 1, we propose the signal adversarial examples generation model as one that minimizes the number of detected targets in signal adversarial example, under the constraint that the perturbation L2-norm is less than a certain proportion of the original signal's time-frequency domain L2-norm. The model is defined as follows: 
\begin{equation}
\begin{matrix}
  \min \text{ }f({{x}_{r}}+{{\delta }_{r}}\ ;\theta ) \\ 
  s.t.\quad {{y}_{r}}=\mathop{f}_{1}({{x}_{r}}) \\ 
  ||{{\delta }_{r}}|{{|}_{2}}\le ||{{y}_{r}}|{{|}_{2}}*{\alpha }' \\
\end{matrix}	
\end{equation}
${{\delta }_{r}}$ represents the adversarial perturbation added to the signal sample ${{x}_{r}}$, resulting in the signal adversarial example $x_{r}^{*}={{x}_{r}}+{{\delta }_{r}}$, $||{{y}_{r}}|{{|}_{2}}$ is the L2-norm of the time-frequency domain signal sample, ${\alpha }'$ is the maximum ratio value of $||{{\delta }_{r}}|{{|}_{2}}$ compared to $||{{y}_{r}}|{{|}_{2}}$, which is determined by the equation ${\alpha }'=\sqrt{\frac{3}{N}}*\alpha $.

\subsection{White-Box Signal Adversarial Attack Method for Signal Detection Network}\label{subsec2}

In Section 2.1, we present the signal adversarial examples generation model for signal detection network, and the design of the loss function. This section primarily discusses the generation method of signal adversarial examples for signal detection network under white-box conditions. 

During the process of generating signal adversarial examples, the application of gradient-based attack methods is obstructed by the time-frequency diagram transformation module, which involves complex-number computations, thereby complicating the backpropagation of gradient information. Our method is converting the time domain adversarial perturbation into time-frequency domain adversarial perturbations. After generating the time-frequency domain adversarial examples, use time-frequency conversion to restore them to time domain signal adversarial examples.

With the loss function of Section 2.1.2, we will use the PGD (Projected Gradient Descent) attack algorithm as an example to elaborate on the process of signal detection network’s signal adversarial examples generation:

\begin{algorithm}
\caption{ Signal Adversarial Examples’ Generation for Signal Detection Network via PGD}\label{algo1}
\begin{algorithmic}[1]
\Statex \hspace*{-\algorithmicindent} \parbox{\textwidth}{\textbf{Input:} Input signal $x$, original detection results $O$, signal detection network $f$, time-frequency diagram conversion module $\mathop{f}_{1}$, object detection network $\mathop{f}_{2}$, loss function L}
\Statex \hspace*{-\algorithmicindent} \parbox{\textwidth}{\textbf{Parameters:}  Iteration number $N$, perturbation amount $eps$, perturbation step size $\varepsilon $, decay factor $d$, maximum perturbation L2-norm ratio $\alpha $} 
\Statex \hspace*{-\algorithmicindent} \parbox{\textwidth}{\textbf{Output:} Signal adversarial example for signal detection network ${x}'$ }
\State $\vec{y}=\mathop{f}_{1}\left( x \right)$, $y(k,m)$ is the signal amplitude matrix of $\vec{y}$, $\theta (k,m)$ is the phase matrix of $\vec{y}$
\State ${{y}_{n}}$ is the result of the n-th iteration, let ${{x}_{0}}=x$, ${{y}_{0}}={{f}_{1}}({{x}_{0}})$, $n=0$
\While{${{f}_{2}}({{y}_{n}})\ne 0$ and $n<N$ }
        \State $g\leftarrow {{\nabla }_{y}}L({{y}_{n}},O)$
        \State $\beta \leftarrow \varepsilon g/{{\left| g \right|}_{2}}$
        \State $\beta \leftarrow Clip2(\beta ,eps)$
        \While{${{\left\| {{y}_{n}}-y \right\|}_{2}}>\alpha {{\left\| y \right\|}_{2}}$}
                \State $\beta \leftarrow d*\beta $
        \EndWhile
        \State $n\leftarrow n+1$, ${{y}_{n}}\leftarrow {{y}_{n-1}}-\beta $
\EndWhile
\State ${y}'={{y}_{n}}$, ${x}'=f_{1}^{-1}({y}',\theta )$
\State \textbf{Return:} ${x}'$
\end{algorithmic}
\end{algorithm}

In Algorithm 1, we use the disappearance of signal targets in the detection results as the loss. The signal sample $x$ is converted to time-frequency matrix $\vec{y}$ through time-frequency conversion. The time-frequency matrix $\vec{y}$ contains signal amplitude $y(k,m)$ and phase $\theta (k,m)$ information, and the phase information $\theta (k,m)$ will be saved; then the gradient-based attack algorithm (PGD is used as an example here) is used to attack the amplitude matrix $y(k,m)$. We use the complete disappearance of the target ${{f}_{2}}({{y}_{n}})=0$ and the number of iteration rounds $n=N$ as the iteration termination conditions, and constrain the energy of the perturbation $\left\| {{y}_{n}}-y \right\|_{2}^{2}\le \alpha \left\| y \right\|_{2}^{2}$. After multiple iterations, the time-frequency domain signal adversarial example ${y}'$ are obtained. Then the time-frequency domain signal adversarial example ${y}'$ combing with the signal phase information $\theta (k,m)$ is transformed into time domain signal adversarial example through inverse short-time Fourier transform ${x}'=f_{1}^{-1}({y}',\theta )$.

Our model can modify the PGD attack algorithm to other gradient-based attack algorithms to generate signal adversarial examples for signal detection network.

\section{Experiment}\label{sec3}

In order to verify the effectiveness of the signal adversarial examples generation model, this chapter mainly introduces the experimental related situations. Firstly, we use the original clean signal to train the signal detection network and count its training results, then we use two gradient-based attack algorithms, FGM and PGD, to generate signal adversarial examples under different energy ratio constraints respectively, and statistically count the attack effect of the signal adversarial examples and compare it with the random noise under the same energy perturbation, and finally we present the waveform diagram of the signal adversarial examples to visually perceive the adversarial perturbations to the signals. It is important to note that there are no definitive metrics for evaluating the effectiveness of adversarial attacks on signal detection network. Hence, in this study, we temporarily employ metrics such as mean Average Precision (mAP), recall, and precision from the domain of object detection tasks to demonstrate the impact of the attacks.

\subsection{Experimental Setting}\label{subsec3}

The signal dataset file is stored in the form of 16-bit integer binary, and a single signal file contains a total of 16*1280000 binary data, which can be converted to 1280000 16-bit integer data.
The signal detection network model consists of a time-frequency diagram conversion module and a YOLOV5 object detection network cascade. When processing the signal data, the time-frequency diagram conversion module reads a single signal file as 16-bit integer data, and then normalizes the read signal data, sets the number of short-time Fourier transform points (STFT) to 2048, the sampling rate to 6400000, the number of overlap points to 48, and uses the Blackman window function. The generated time-frequency diagram is set to be a grey scale graph, and the range of grey scale values is set to be from 0 to 255. The labels in the dataset are originally the frequency information, time information and category information of the signals, and the time-frequency information of the labels is converted into the coordinate information of the signal targets in the time-frequency diagrams. We then use the generated time-frequency diagrams and labels to construct dataset and train the YOLOV5 network. The mAP of the trained YOLOV5 detection network on the signal detection dataset can reach 0.815, the recall rate is 0.781, and the precision rate is 0.811.

\subsection{Error in Signal Time-Frequency Inversion and Selection of Perturbation L2-norm Ratios}\label{subsec3}

In this section, we do time-frequency inversion of the original signal time-frequency diagrams to count the error. Subsequently, we use two attack algorithms, FGM and PGD, to generate signal adversarial examples, and count the time domain perturbation L2-norm ratio distributions to assist in selecting the appropriate signal time-frequency domain perturbation ratios to implement the attack.
The experiments respectively use 1\%, 2\%, 5\% and 10\% as the ratio of the signal time-frequency domain perturbation L2-norm to the clean signal time-frequency domain L2-norm. Since the short-time Fourier transform and inversion process of the signal are affected by the window function, overlap processing, numerical accuracy, signal boundary effects, etc., which can make the clean signal time-frequency domain samples have errors in the process of recovering, we have likewise conducted experiments and statistics on the time-frequency inversion process of the clean signal as a reference. The experimental results are shown in Table 1.

\begin{table}[h]
\caption{FGM and PGD signal adversarial examples time-frequency domain and time domain L2-perturbation ratio}\label{tab1}%
\begin{tabular}{@{}p{2.3cm} p{2.3cm} p{2.3cm} p{2.3cm} p{2.3cm}@{}}
\toprule
Attack Method & Ratio of time-frequency domain perturbation & Average ratio of time domain perturbation & Maximum ratio of time domain perturbation & Minimum ratio of time domain perturbation\\
\midrule
None    & 0\%    & 4.692\%    & 8.761\%    & 3.523\% \\
FGM    & 1\%    & 7.61\%    & 13.59\%    & 6.197\% \\
FGM    & 2\%    & 9.947\%    & 24.64\%    & 6.924\% \\
FGM    & 5\%    & 22.07\%    & 82.86\%    & 11.61\% \\
FGM    & 10\%    & 52.44\%    & 138.97\%    & 28.85\% \\
PGD    & 1\%    & 7.545\%    & 12.34\%    & 6.11\% \\
PGD    & 2\%    & 9.661\%    & 17.48\%    & 7.047\% \\
PGD    & 5\%    & 21.02\%    & 41.78\%    & 11.40\% \\
PGD    & 10\%    & 52.11\%    & 123.16\%    & 27.36\% \\
\botrule
\end{tabular}
\end{table}

Under the two attack methods of FGM and PGD, we obtain the proportions of time domain adversarial perturbations corresponding to the different proportions’ time-frequency domain perturbations. It can be observed that the proportion of the signal time domain perturbation L2-norm is positively correlated with the proportion of the time-frequency domain perturbation L2-norm, and as the proportion of the time-frequency domain perturbation increases, the proportion of the signal time domain perturbation increases as well. The average error value after inverse transformation of the original signal samples is about 4.692\%. The average proportion in the time domain is less than 10\% under the PGD perturbation of a 2\% proportion of the time-frequency domain, with a maximum of 17.48\%, which, if converted to the energy is then 1\% and 3\%, which can be roughly converted to -20 dB and -15 dB. Under this condition, the perturbation is usually considered not to affect the communication properly. Therefore, in the subsequent experiments, 2\% and 1\% were used as the time-frequency domain perturbation ratios.

\subsection{Signal Adversarial Attack Results}\label{subsec3}

In Section 3.2, by analyzing the time domain perturbations’ energy under different proportions’ time-frequency domain perturbations, we select 2\% and 1\% as the l2 perturbation proportions of the time-frequency domain perturbation and generate signal adversarial examples under this constraint. To serve as a comparison, we add random noise with the same signal time-frequency domain l2 perturbation ratio to the clean signal sample and statistically the detection results. The experimental results are as follows:

\begin{table}[h]
\caption{Detection results of different signal samples}\label{tab1}%
\begin{tabular}{@{}p{2cm} p{2cm} p{2cm} p{2cm}@{}}
\toprule
Sample type &	mAP &	Recall &	Precision\\
\midrule
Sample    & 0.815    & 0.781    & 0.811\\
RN\_0.01    & 0.816    & 0.782    & 0.811\\
RN\_0.02    & 0.82    & 0.783    & 0.818\\
RN\_0.05    & 0.831    & 0.785    & 0.84\\
RN\_0.1    & 0.816    & 0.754    & 0.826\\
FGM\_0.01    & 0.645    & 0.624    & 0.636\\
FGM\_0.02    & 0.545    & 0.539    & 0.519\\
PGD\_0.01    & 0.701    & 0.673    & 0.626\\
PGD\_0.02    & 0.534    & 0.534    & 0.507\\
\botrule
\end{tabular}
\end{table}

As shown in Table 2, the mAP values, recall and precision are essentially unchanged under 1\%-10\% proportion of random noise (RN) perturbation in comparison to clean samples. With our model via FGM method, 2\% time-frequency domain L2-norm perturbation caused the mAP values of the signal detection network to decrease by 27\%, recall rates by 24.2\%, and precision rates by 29.2\%. Under the same constraints, when employing the PGD method, the mAP values decrease by 28.1\%, recall rates by 24.7\%, and precision rates by 30.4\%. The three key metrics of the signal detection network were also significantly reduced under 1\% time-frequency domain L2-norm perturbation. Unlike random noise, our signal adversarial examples generation model has a disturbing effect on the signal detection network, resulting in a significant decrease in the performance metrics.
In order to show more visually the changes in the signal adversarial examples compared to the original signals, we plot the waveforms as follows: 

\begin{figure}[htbp]
\centering
\subfigure[]
{
    \begin{minipage}[b]{.48\linewidth}
        \centering
        \includegraphics[scale=0.4]{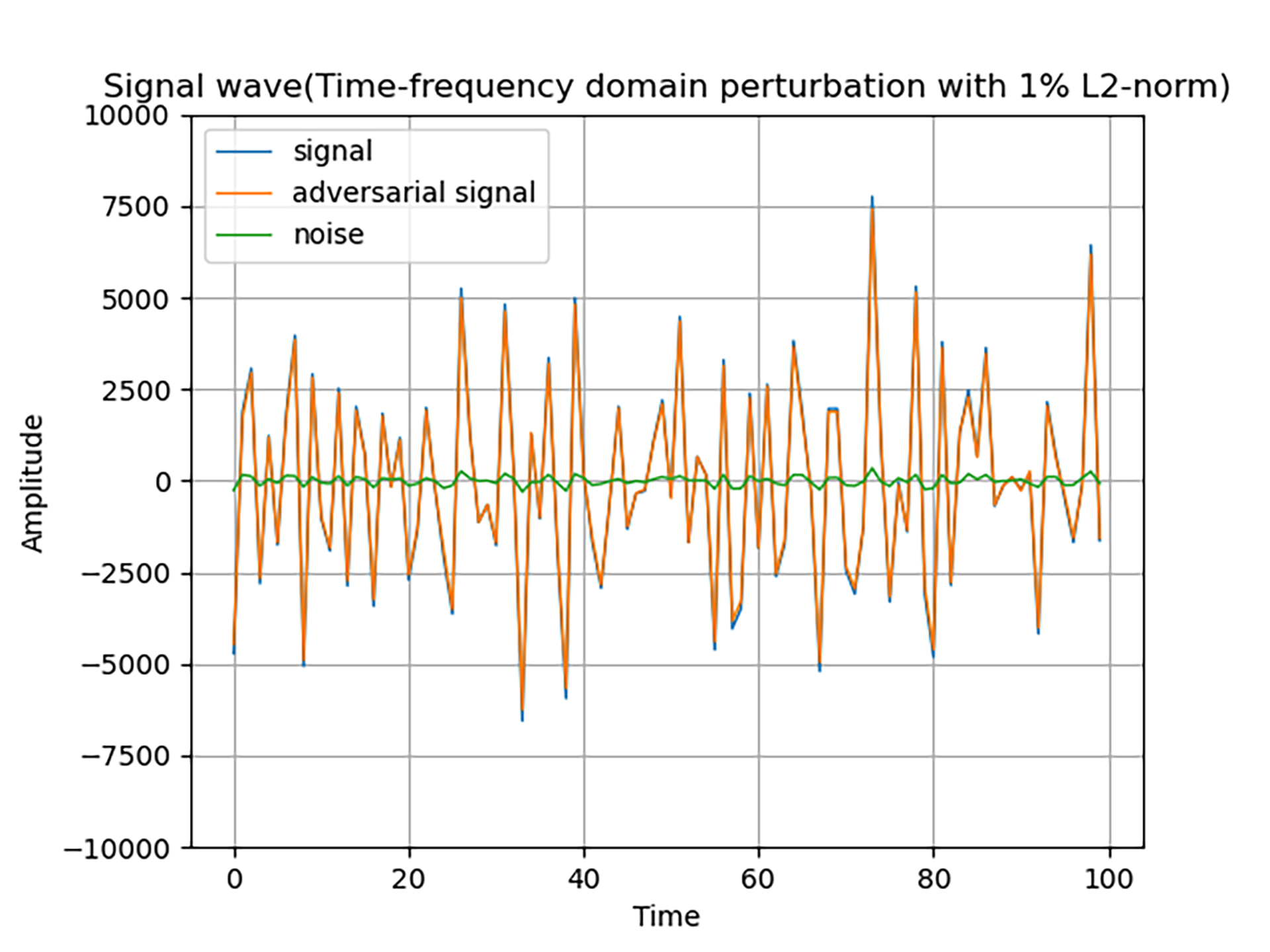}
    \end{minipage}
}
\subfigure[]
{
 	\begin{minipage}[b]{.48\linewidth}
        \centering
        \includegraphics[scale=0.4]{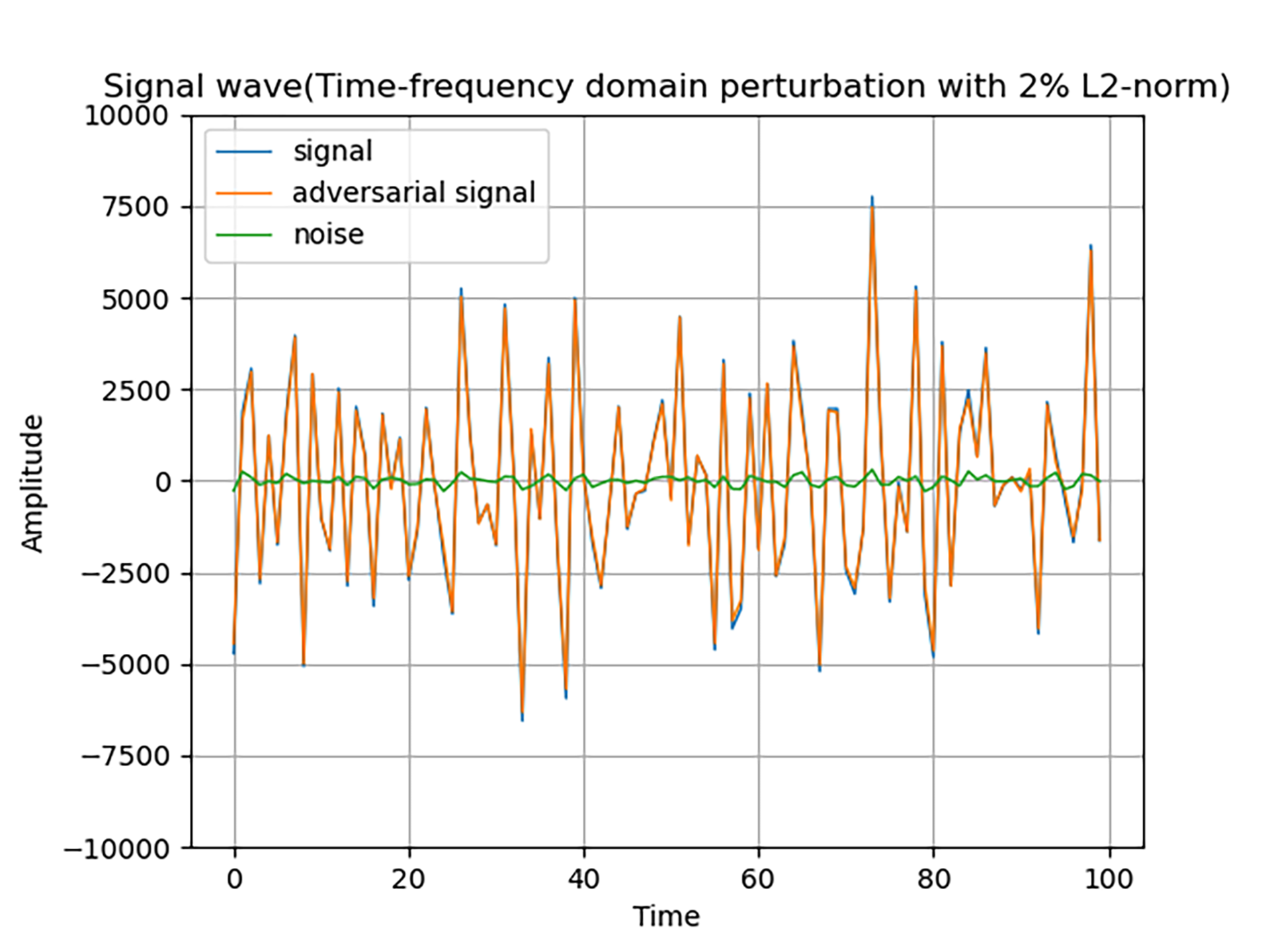}
    \end{minipage}
}
\caption{\centering{Signal time-frequency diagram and detection results}}\label{fig4}
\end{figure}

In Figure 4, subfigure (a) shows the 1\% time-frequency domain L2-norm perturbation, subfigure (b) shows the 2\% time-frequency domain L2-norm perturbation, the blue curve is the original clean signal sample, the yellow curve is the signal adversarial example, and the green curve is the added adversarial perturbation. From the figure, it can be observed that the added perturbations have minimal effect on the signal waveforms under 1\% and 2\% of the L2-norm constraints, which are almost consistent with the original waveforms. It is worth noting that the three key metrics decrease more for the 2\% constraint, which suggests that choosing the 2\% L2-norm constraint will be more effective in implementing the attack as it achieves a more substantial metric reduction while maintaining the waveform similarity.

\section{Conclusion}\label{sec4}

This paper proposes a signal adversarial examples generation model for signal detection network, and gives a signal adversarial examples generation method via white-box attack. Drawing on evaluation metrics from object detection, under the 2\% time-frequency domain perturbation constraint (corresponding to the time domain SNR of -20dB to -15dB), the mAP values decrease by 28.1\%, the recall decreases by 24.7\%, and the precision decreases by 30.4\% compared with the clean samples, while the random noise of the same intensity does not effectively interfere with the signal detection network. From the waveform diagram, it shows that signal adversarial examples have only minor changes in the waveform compared with the original signal, which is not easy to be noticed. Future research will investigate the evaluation metrics for adversarial examples in signal detection networks, as well as adversarial attacks under various scenarios.

\begin{appendices}

\section{Proof for Theorem 1}\label{secA1}

Here, we provide proof of Theorem 1.

Let $x[n]$ be the input signal sequence, after short-time Fourier transform, time-frequency matrix $y(k,m)=\mathop{f}_{1}(x[n])$ is obtained. The transformation formula is as follows:
\begin{equation}
y(k,m)=\sum\limits_{n=0}^{N-1}{x[n]w[n-mR]{{e}^{\frac{-j2\pi kn}{N}}}}
\end{equation}
where $k$ is the frequency index, $m$ is the time index, $N$ is the length and $R$ is the moving step size of the time window $w$, and $y(k,m)$ is a complex matrix containing the amplitude and phase information of signal $x[n]$. Adding perturbation $t(k,m)$ to the amplitude value of $y(k,m)$, the resulting new signal time-frequency matrix ${{y}_{1}}(k,m)$ is:
\begin{equation}
{{y}_{1}}(k,m)=\left( \left| y(k,m) \right|+t(k,m) \right){{e}^{j\angle y(k,m)}}
\end{equation}

The L2-norm of the perturbation added to the amplitude in the time-frequency domain is given by: 
\begin{equation}
||{{\beta }_{r}}|{{|}_{2}}=\sqrt{\sum\limits_{k=0}^{N-1}{\sum\limits_{m=-\infty }^{\infty }{{{t}^{2}}(k,m)}}}
\end{equation}
Here, $t(k,m)$ represents the perturbation adding to time-frequency matrix. This equation quantifies the magnitude of the perturbation in the time-frequency domain. 

The time-frequency domain signal adversarial example ${{y}_{1}}(k,m)$ is restored to the time domain ${{x}_{1}}[n]$ through the inverse short-time Fourier transform. Then the corresponding L2-norm of time domain perturbation ${{\delta }_{r}}$ is:
\begin{equation}
||{{\delta }_{r}}|{{|}_{2}}={{\left\| {{x}_{1}}-x \right\|}_{2}}
\end{equation}

The inverse short-time Fourier transformation process of ${{y}_{1}}(k,m)$ is: 
\begin{align}
{{x}_{1}}[n] &= \frac{1}{N}\sum\limits_{m=-\infty }^{\infty }{\sum\limits_{k=0}^{N-1}{{{y}_{1}}(k,m)w(n-mR){{e}^{\frac{j2\pi kn}{N}}}}} \nonumber \\
&= \frac{1}{N}\sum\limits_{m=-\infty }^{\infty }{\sum\limits_{k=0}^{N-1}{\left( \left| y(k,m) \right|+t(k,m) \right){{e}^{j\angle y(k,m)}}w(n-mR){{e}^{\frac{j2\pi kn}{N}}}}}
\end{align}

The inverse short-time Fourier transformation process between the original signal $x[n]$ and the time-frequency matrix $y(k,m)$ is:
\begin{align}
x[n] &= \frac{1}{N}\sum\limits_{m=-\infty }^{\infty }{\sum\limits_{k=0}^{N-1}{y(k,m)w(n-mR){{e}^{\frac{j2\pi kn}{N}}}}} \nonumber \\
&= \frac{1}{N}\sum\limits_{m=-\infty }^{\infty }{\sum\limits_{k=0}^{N-1}{\left| y(k,m) \right|{{e}^{j\angle y(k,m)}}w(n-mR){{e}^{\frac{j2\pi kn}{N}}}}}
\end{align}

Then the corresponding L2-norm of time domain perturbation ${{\delta }_{r}}$ is:
\begin{align}
||{{\delta }_{r}}|{{|}_{2}} &= {{\left\| {{x}_{1}}-x \right\|}_{2}}= \sqrt{\sum\limits_{n=0}^{N-1}{{{\left| {{x}_{1}}(n)-x(n) \right|}^{2}}}} \nonumber \\
&= \sqrt{\sum\limits_{n=0}^{N-1}{{{\left| \frac{1}{N}\sum\limits_{m=-\infty }^{\infty }{\sum\limits_{k=0}^{N-1}{\left| t(k,m) \right|{{e}^{j\angle y(k,m)}}{{e}^{j2\pi kn/N}}}} \right|}^{2}}}}
\end{align}
This equation quantifies the magnitude of the perturbation in the time domain. 

Next, we will prove the inequality relationship between the sum of squares of n numbers and the square of the magnitude of the resultant vector obtained by vector addition, where each vector is of a magnitude corresponding to these n numbers and with random angles.

Let $A$ be the sum of squares of $N$ elements ${{a}_{i}}$, ${{\vec{v}}_{i}}$ be the vector of a magnitude corresponding to a and with random angle, and $\vec{V}$ be the resultant vector of the sum of vectors ${{\vec{v}}_{i}}$:
\begin{equation}
\begin{matrix}
  A=\sum\limits_{i=1}^{N}{a_{i}^{2}} \\ 
  {{{\vec{v}}}_{i}}={{a}_{i}}{{e}^{j{{\theta }_{i}}}} \\ 
  \vec{V}=\sum\limits_{i=1}^{N}{{{{\vec{v}}}_{i}}}=\sum\limits_{i=1}^{N}{{{a}_{i}}{{e}^{j{{\theta }_{i}}}}} \\ 
\end{matrix}
\end{equation}

Then, the magnitude of $\vec{V}$ is given by:
\begin{equation}
{{\left| {\vec{V}} \right|}^{2}}=\vec{V}\cdot \vec{V}=\left( \sum\limits_{i=1}^{N}{{{{\vec{v}}}_{i}}} \right)\cdot \left( \sum\limits_{j=1}^{N}{{{{\vec{v}}}_{j}}} \right)=\sum\limits_{i=1}^{N}{{{{\vec{v}}}_{i}}}\cdot {{\vec{v}}_{i}}+\sum\limits_{i\ne j}{{{{\vec{v}}}_{i}}}\cdot {{\vec{v}}_{j}}=\sum\limits_{i=1}^{N}{a_{i}^{2}}+\sum\limits_{i\ne j}{{{{\vec{v}}}_{i}}}\cdot {{\vec{v}}_{j}}
\end{equation}

Based on the inequality relationship of vector magnitudes, it can be derived that: 
\begin{equation}
\sum\limits_{i\ne j}{{{{\vec{v}}}_{i}}}\cdot {{\vec{v}}_{j}}\le \sum\limits_{i\ne j}{\left| {{{\vec{v}}}_{i}} \right|}\cdot \left| {{{\vec{v}}}_{j}} \right|\le 2\sum\limits_{i=1}^{N}{{{\left| {{{\vec{v}}}_{i}} \right|}^{2}}}=2\sum\limits_{i=1}^{N}{a_{i}^{2}}
\end{equation}

From the above, we can obtain:
\begin{equation}
{{\left| {\vec{V}} \right|}^{2}}=\vec{V}\cdot \vec{V}=\sum\limits_{i=1}^{N}{a_{i}^{2}}+\sum\limits_{i\ne j}{{{{\vec{v}}}_{i}}}\cdot {{\vec{v}}_{j}}\le 3\sum\limits_{i=1}^{N}{a_{i}^{2}}
\end{equation}

Based on the derived inequality relationship, we can conclude that:
\begin{equation}
{{\left| \sum\limits_{m=-\infty }^{\infty }{\sum\limits_{k=0}^{N-1}{\left| t(k,m) \right|{{e}^{j\angle y(k,m)}}{{e}^{j2\pi kn/N}}}} \right|}^{2}}\le 3\sum\limits_{k=0}^{N-1}{\sum\limits_{m=-\infty }^{\infty }{{{t}^{2}}(k,m)}}=3||{{\beta }_{r}}||_{2}^{2}
\end{equation}

The inequality relationship between the L2-norms of ${{\delta }_{r}}$ and ${{\beta }_{r}}$ is as follows:
\begin{align}
||{{\delta }_{r}}|{{|}_{2}} &= \sqrt{\sum\limits_{n=0}^{N-1}{{{\left| \frac{1}{N}\sum\limits_{m=-\infty }^{\infty }{\sum\limits_{k=0}^{N-1}{\left| t(k,m) \right|{{e}^{j\angle y(k,m)}}{{e}^{j2\pi kn/N}}}} \right|}^{2}}}} \nonumber \\
&\le \sqrt{\frac{3}{N}\sum\limits_{k=0}^{N-1}{\sum\limits_{m=-\infty }^{\infty }{{{t}^{2}}(k,m)}}} = \sqrt{\frac{3}{N}}||{{\beta }_{r}}|{{|}_{2}}
\end{align}

Thus, $||{{\delta }_{r}}|{{|}_{2}}$ is less than or equal to $\sqrt{\frac{3}{N}}||{{\beta }_{r}}|{{|}_{2}}$.




\end{appendices}


\bibliography{sn-bibliography}% common bib file

\end{document}